\documentclass{article}
\usepackage{amsmath}
\usepackage{xcolor}
\usepackage[
  a4paper,
  margin=1in,
  headsep=15pt,
  left=0.75in,
  right=0.75in
]{geometry}
\usepackage{fancyhdr}
\usepackage{textcomp}

\usepackage{ graphicx}
\graphicspath{ {./images/} }

\usepackage{hyperref}
\usepackage{amssymb}

\usepackage{float}

\fancypagestyle{firststyle}
{

}
\begin{document}
\vspace{-2mm}
\hrule
\begin{center}\large{\textbf{Ingredient Extraction from Text in the Recipe Domain.}}
\end{center}
\begin{center}\large{Arkin Dharawat -- adharawat@umass.edu\\ Chris Doan -- chrdoan@umass.edu\\}
\end{center}

\vspace{5em}
\section{Abstract}
In recent years, there has been an increase in the number of devices with virtual assistants (e.g: Siri, Google Home, Alexa) in our living rooms and kitchens. As a result of this, these devices receive several queries about recipes. All these queries will contain terms relating to a "recipe-domain" i.e: they will contain dish-names, ingredients, cooking times, dietary preferences etc. Extracting these recipe-relevant aspects from the query thus becomes important when it comes to addressing the user's information need. Our project focuses on extracting ingredients from such plain-text user utterances. Our best performing model was a fine-tuned BERT which achieved an F1-score of $95.01$. We have released all our code in a GitHub repository \footnote{https://github.com/ArkinDharawat/ingredient\_extraction}.

\section{Introduction}
Exploration of new recipes and new cuisines is an activity engaged in by people ever since search engines have become ubiquitous. A report from Google shows that "59\% of 25- to 34-year-olds cook with either their smartphones or tablets handy" \footnote{https://www.thinkwithgoogle.com/future-of-marketing/digital-transformation/cooking-trends-among-millennials/}. A user might have several reasons behind doing this which can range from simple curiosity, an acitvity with friends or to satisfying hunger. Given the fact that virtual assistants are present in almost all our devices, it is not quite far-fetched to assume that user's will want to address their recipe-related information need through these agents. To address a user's information need in this restricted domain it is important to understand the vocabulary and extract relevant information from the query. \\
\\
Extracting aspects from user queries has been an active research area in the fields of Information Retrieval (IR) and Natural Language Processing (NLP). E.g: 
Extraction of clinical entities from user queries \cite{zhao2019extracting}. Earlier approaches in extractions take advantage of the query or text conforming to a certain structure or used well-known unsupervised methods \cite{jain-pennacchiotti-2010-open}.
Virtual assistants at home take the user utterance and convert it to text using Automatic Speech Recognition (ASR). These queries made by the user are short and do not conform to any particular structure, they may also have minor spelling error based on the performance of the ASR model.\\
\\
We choose to focus on  ingredients since they can't be detected by simple hard-coding (e.g: make an exhaustive list which can done in the case of dietary preferences) or use a regex that conforms to some structure (e.g: in the case of cooking times.). Our task can be described as extraction of ingredients from a raw user query in the recipe domain. Our approach to this problem is framed as an extraction of entities (ingredients), commonly also known as Named Entity recognition (NER). We use NER techniques and models where the ingredients are our desired entities and the rest of the text is to be ignored. Thus our goal is the correct extraction of ingredients from a user text. \\
To achieve this we have taken the following steps, all which will be covered in detail later:
\begin{enumerate}
    \item Created and annotated our own dataset by using publicly available data on recipes.
    \item Pre-processed and cleaned available datasets. Combined our dataset with other similarly annotated data.
    \item Trained NER models: one using traditional NLP techniques and another using deep-learning and transformers. 
    \item Generate our own test data and test the performance of our models on unseen real-world data. 
\end{enumerate}
Finally, We evaluate the performance of our models and analyse areas of improvements for interested researchers. 

\section{Background/Related Work}
We looked into several techniques of information extraction for food/recipes.

In the FoodIE paper \cite{inproceedings}, named entity recognition is performed by building on and improving semantic tagging using the UCREL Semantic Analysis System (USAS).  The USAS is an automatic semantic tagger which has many categories related to food. FoodIE uses semantic tags and its own rule based system to extract food phrase chunks from recipe data. The model was evaluated on 200 food recipes and achieved a F1 score of 0.9593 on food entity tagging.  Most errors occur with false negatives classifications regarding brand name used in recipes, colloquial names, and dual part-of-speech of certain words.

A Stanford paper regarding information extraction from recipes was also reviewed\cite{IEfromrecipe} where a maximum entropy Markov model was used to extract many food labels like ingredients, utensil, etc from recipe data. The model evaluated well for ingredients but did not do well for utensils.

Although this area has been explored, what we discovered was that several of the datasets were trained on recipe text and relied on some structure of the text. E.g: FoodIE is a rule-based method that relies on the ingredients having a certain POS tag to determine if a word could be an ingredient.

\section{Approach}
Our problem statement can be formally defined as follows: Given a dataset $D$ where each element is tuple that can be defined as $(t_i, L_i)$. $t_i$ is the recipe text and $L_i$ are the annotations in the form of $[(s_0, e_0) \cdots (s_{n_i}, e_{n_i})]$ where $s_i$ indicates the start of the span, $e_i$ indicates the end of the span and $n_i$ are the total number of annotations for $t_i$. Our model $f$ takes in $t_i$ only and outputs the list of start-end spans for the labels.
\subsection{Dataset}
As mentioned in our proposal, we planned to use two datasets. Currently, we are using the following publicly-available datasets:
\begin{enumerate}
    \item \textbf{FoodBase} \cite{10.1093/database/baz121}: This dataset consists of recipes with their text and annotated ingredients. It contains 1K human annotated recipes and ~12K annotated by a heuristic. 
    \item \textbf{Food.com}: A publicly available dataset of Food.com available on Kaggle \footnote{\url{https://www.kaggle.com/shuyangli94/food-com-recipes-and-user-interactions}}. Each row is a recipe with number of steps, text of the steps, ingredients etc. We will use this dataset to generate text with the ingredients labelled by their indices. 
\end{enumerate}
We are using these for training and evaluation with a train/eval/test split of $80/10/10$.\\
We used the NLTK \cite{LoperBird02} to do the text pre-processing. Our pre-processing for FoodBase required us to parse and tokenize the XML data and then convert it into a JSON format suitable for SpaCy. However, this wasn't a streamlined process since we encountered several issues: 
\begin{enumerate}
    \item Overlapping spans: Several of the annotations were overlapping e.g: 'cream cheese' sometimes had two annotations one was for 'cream' and another was for 'cream cheese'.
    \item Duplicate annotations: Several times ingredients were annotated twice-thrice i.e: the same span start-end pairs occurred more than once. 
\end{enumerate}

In the case for the Food.com dataset we only had the recipe-text and list of ingredients. We had to manually annotate some of the data together. To achieve this we used a free annotation tool, Doccano \cite{doccano}. We setup a local docker container where we uploaded a cleaned and formatted version of our dataset. We each took turns annotating as much text as we could.\\
Here are the statistics for each of our datasets:
\begin{table}[H]
\centering
\begin{tabular}{|l|l|l|l|}
\hline
Dataset  & Total Rows & Total labels \\ \hline
Foodbase & 22669 & 271565 \\ \hline
Food.com & 106 & 1100 \\ \hline
\end{tabular}
\caption{Dataset statistics}
\end{table} 
The following are top 5 ingredients for Foodbase:
\begin{table}[H]
\centering
\begin{tabular}{|l|l|}
\hline
Ingredient  & Count \\ \hline
Sugar &  10580\\ \hline
Flour &  9355 \\ \hline
Butter &  8437 \\ \hline
Salt &  8342 \\ \hline
Eggs &  7114 \\ \hline
\end{tabular}
\caption{Foodbase ingredients}
\end{table} 
Top 5 ingredients for Food.com:
\begin{table}[H]
\centering
\begin{tabular}{|l|l|}
\hline
Ingredient  & Count \\ \hline
Butter &  53\\ \hline
Salt &  42 \\ \hline
Water &  36 \\ \hline
Cheese &  34 \\ \hline
Flour &  33 \\ \hline
\end{tabular}
\caption{Food.com ingredients}
\end{table} 
As we can notice some of our most common ingredients are \textbf{butter}, \textbf{salt}, \textbf{flour} and \textbf{sugar} which makes sense since these would occur in a lot of dishes.  For ease of reference we'll refer to the combination of \textbf{FoodBase} and \textbf{Food.com} as FoodCombined. Additionally, we also had to convert the rows in FoodCombined to specific formats for SpaCy and BERT. Our SpaCy model used a simple json with the accompanied spans the BERT model used a csv with "Insider-Outside-Beginning" (IOB) format. This differentiatde from the SpaCy that took a single INGREDIENT entity tag while for the BERT model we needed two specific tags for ingredients: "B-ING and I-ING". This allows the model to learn chunks of tokens as ingredients like "ground beef". "B-ING" indicates the start of a new chunk of ingredients and "I-ING" indicates the inside of a chunk. In the "ground beef" example beef would be tagged as "B-ING" and beef would be tagged as "I-ING". This add some complexity to our training since the chunking of ingredients allows our model to interpret more than single token-ed ingredients but can be susceptible to over fitting to our trainings data and the chunks that exist in the corpus.\\
\\
It is always good practice to evaluate your model on unseen data which is usually refereed to as the test set. However, in our case there was a small discrepancy between the test set and what we expected in the real-world. Our dataset consisted of annotated recipes, the text for each row consisted of a few sentences and had more than a single ingredient. The user utterances that are made to the virtual assistants usually are only a single sentence along with one to three ingredients mentioned.\\
\\
To generate a dataset that was closer to the real world, we decided to use Checklist \cite{ribeiro2020accuracy}, an NLP testing framework. It is important to note that so far Checklist has only been used for testing text classification models by generating and perturbing text based on certain templates. We took advantage of it's generative functionality to generate $500$ test cases. These test cases were generated by a variety of templates we came up with, these include:
\begin{enumerate}
    \item "I would like a dish with \{ing\}", 
    \item "Show me \{ing\} recipes", 
    \item "Can I see recipes with \{ing\}",
    \item "I want \{ing\} in my recipe"
\end{enumerate}
The "\{ing\}" in these templates is replaced by the ingredient which we sample from a large ingredient dataset from Kaggle, called Recipe Ingredients Dataset \footnote{https://www.kaggle.com/kaggle/recipe-ingredients-dataset}. Some generated examples are: "Can I see recipes with bacon?", "I would like a dish with peanut oil" and "I want soy sauce in my recipe" etc. For each of these examples we had to manualy calculate the span for each ingredient based on the length of the template and placement of  "\{ing\}".\\
We wanted to generate data with two to three ingredients as well however the generation of the spans became too complicated and error-prone. This is something else we would investigate in the future. 
\subsection{Model}
\subsubsection{SpaCy}
We are working with SpaCy, a free open-source industrial-strength natural language processing library for Python. SpaCy allows for us to use their model framework on custom data and for our use case we are using their NER model framework to correctly identify ingredients in the training recipe data.

SpaCy's NER model is based on the transition-based parser model and the name of its model is "TransitionBasedParser". Transition-based parsers go through a document and creates a word to word relationship graph of each word in the document. This parsing and relationship graphing allows for better understanding of the semantic reasoning and structure of the document \cite{spacy}. The parser begins with an initial state of parsing the document alongside the buffer of all the tokens in the document, and the state is defined as the current set of relationships the parsed words of the document have to one another. The transition based parser then chooses the best scoring state transition operation at each step of the algorithm. State transitions include reading the next token in the buffer, and assigning relationships to words in the parser. Scoring is done by mapping the states to features and taking the dot product with a learn-able weight matrix and summing them up. 
For SpaCy's TransitionBasedParser actions assign tags to words in a state at each step (ie: puts a word into a specific category of entities) \cite{spacy}. SpaCy's implementation also allows for the use of a feed-forward neural network that takes the state representation from the document and calculates the score using the feed forward net.

We pass in our ingredient-annotated recipe training data to train the TransitionBasedParser. From our training step we have a custom entity extractor for our intended problem goal. We can then use this model to evaluate and extract entities from testing data and real samples. The SpaCy model was trained and validated on splits of the FoodCombined dataset with varying hyper-parameter settings. Due to the long training time of the training data, small changes between training sessions were made. The final hyper-parameters included a learning rate of 0.001, a hidden layer width of 64, and batch size equal to 500.

\subsubsection{BERT}

\begin{figure}[h!]
\centering
\includegraphics[width=.5\linewidth]{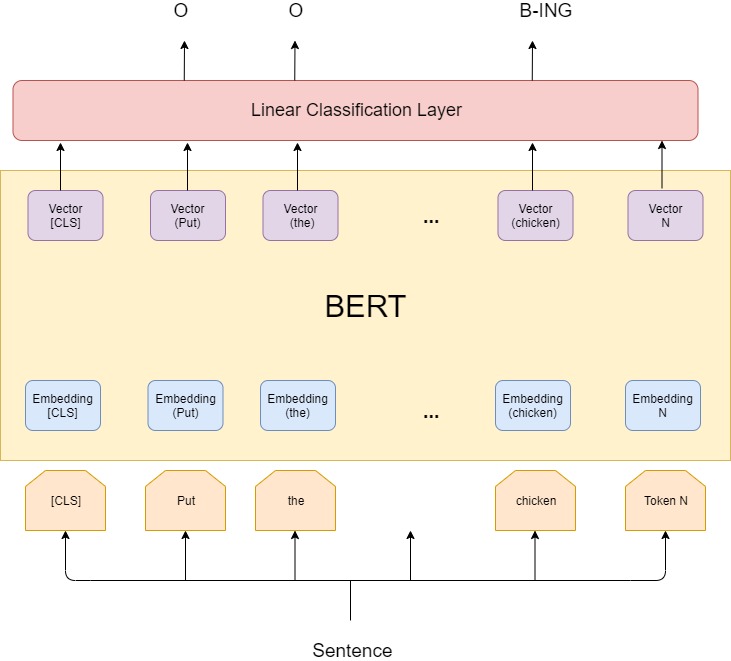}
\end{figure}

The Bidirectional Encoder Representations from Transformers (BERT) is a paper and model proposed by Google AI \cite{DBLP:journals/corr/abs-1810-04805}. BERT is a transformer model that uses bidirectional training as opposed to left-to-right training or a right-to-left training that might occur in an Long Short-term Memory (LSTM) network.

A Transformer is a natural language model that focuses on a special mechanism called attention to capture important information in text data \cite{transformer}. Transformer’s use an encoder that takes input text as an input and generates embeddings of each word with attention based  and positional based vectors. Attention based vectors for each word allow for the model to learn what other words in the same input text are important to that specific word. Using the context from position embeddings and attention embeddings the encoder’s output is used as input for the decoder’s input one-by-one. 

The BERT model is similar to the transformer but it uses only the encoder as its language model alongside some changes to the vanilla transformer to get a bidirectional language model. BERT implements a masked-language model to hide k percent of the input to the encoder in order to guess the values of the missing words and  Next Sentence Prediction pairs sentences together in training to allow for the BERT language model to learn long-term dependencies across sentences. BERT can be used for entity recognition by processing the outputs of the model into a linear layer to classify the labels for each word token \cite{DBLP:journals/corr/abs-1810-04805}.

We use the SimpleTransformer library to use its BERT model. The BERT model was trained using various hyper-parameters that were tuned between training sessions. The final model trained on the entire FoodCombined corpus had the following parameters: learning rate set to $5 \times 10^{-5}$ and batch size equal to 32 in 3 epochs.

\section{Experiment}
\subsection{Process}
The goal of the research is to gather a large corpus of recipe data annotated with variations of "ingredient" tags in order to combine them and use models trained on our corpus to generalize and perform ingredient/food extraction from voice assistant plain-text queries. Our experimentation included using SpaCy's transition based model as a baseline and then to use BERT's transformer based model architecture on our training and evaluating task. The FoodCombined was shuffled and split into a 80:20 train test validation split for training and validation purposes only. Testing on both models used a held out test set of the Checklist created data. 

\subsection{Evaluation Metric}
To evaluate the accuracy of our model using the average correct guess over total word count was not suitable for the data since there are far more non-ingredient words in the dataset than there are ingredient words. This causes an imbalance in the dataset and the best way to measure the performance of our model would be through precision, recall, and F1 score. These metrics use true positives (words that are ingredients from the ground truth are marked as ingredients correctly), false positives (words that are not words are incorrectly labelled as ingredients), false negatives (ingredients from the ground truth are not marked as ingredients), and true negatives. 

Precision is the fraction of true positives over the total number of positives. In our problems scope a true positive would be correctly identifying a word as an ingredient when it is an ingredient.

\begin{equation}
Precision = \frac{TruePositives}{TruePositives + FalsePositives}
\end{equation}

Recall is the fraction of true positives over the sum of true positives and false negatives. In the scope of our problem, this describes the ratio of how many words we correctly label as ingredients out of the total number of words marked as ingredients in the ground truth.

\begin{equation}
Recall = \frac{TruePositives}{TruePositives + FalseNegatives}
\end{equation}

F1 score is the harmonic mean of precision and recall. F1 score helps determine the accuracy of the model when the distribution of ingredient labels and not ingredient labels are imbalanced.

\begin{equation}
F_{1} = 2*\frac{Recall*Precision}{Recall + Precision}
\end{equation}

\subsection{Results: Metrics}
First we trained the SpaCy model on 2000 steps of batch size equal to 500. We then evaluated our trained model on the validation and test splits of data. In Table 4 and 5 you can see the SpaCy NER model achieves a validation F1 score of 96.53\%. Its Recall and Precision scores are also similarly at 96.69\% and 96.67\% respectively. SpaCy's performance on the held-out test set achieved only a F1 score of 55.97\% with lower Recall score of 49.20\% and Precision of 64.91\%. The lower Recall on the test set indicates that the model was not extracting ingredient tags from words that should have been ingredients. The overall accuracy of the test set was very low compared to the validation set. 

\begin{figure}[h!]
\centering
\caption{BERT Training Loss over 3 Epochs}
\includegraphics[width=.5\linewidth]{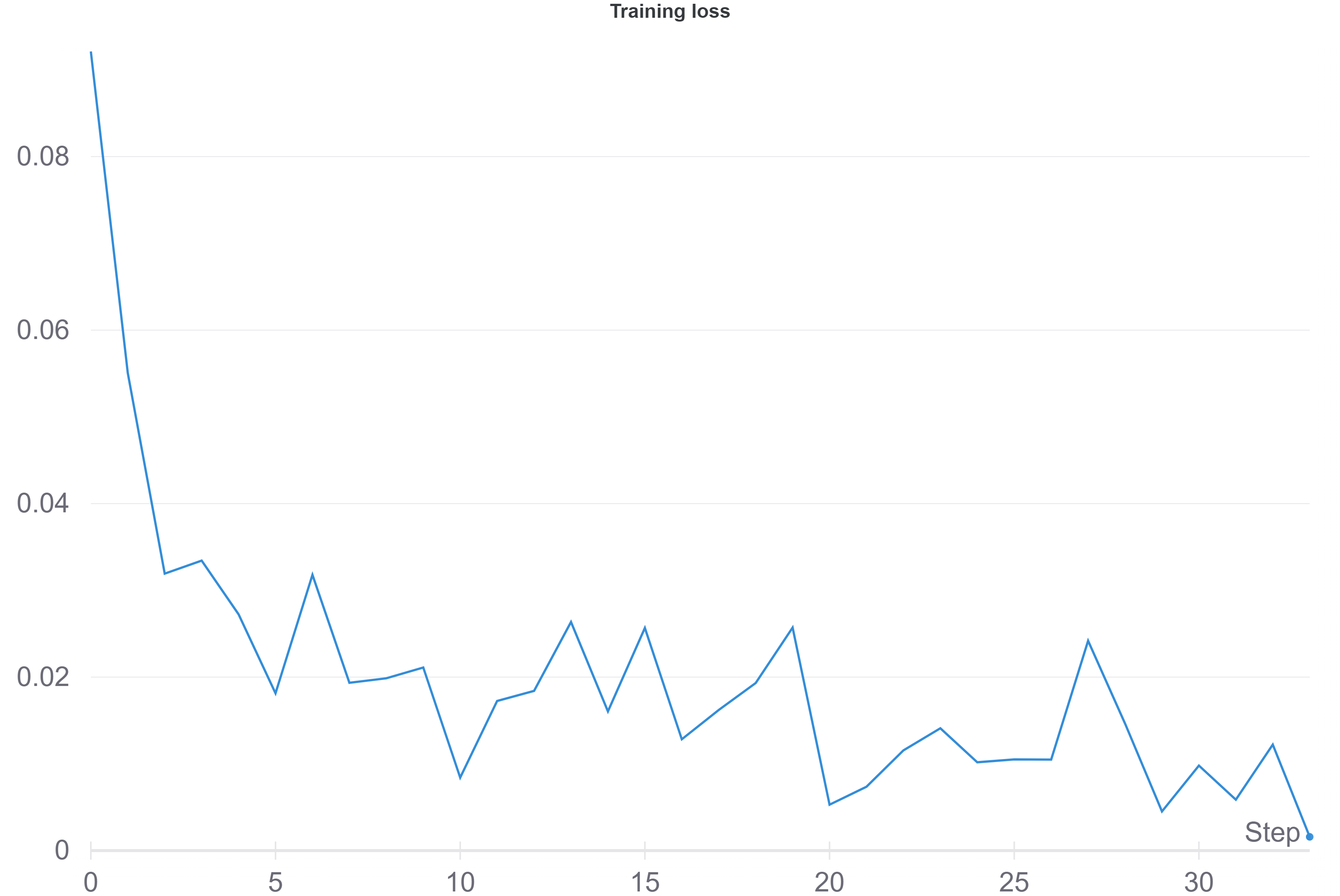}
\end{figure}

To use the power of deep neural networks and transformers we use the BERT model architecture to help us extract ingredient data from textual data. Similar to the SpaCy training and evaluation, we ran BERT for 3 epochs (about 33 steps) on batch size of 32. In Figure 2. we have plotted the training loss over the 3 epochs. The loss decreased exponentially on the first 5-7 steps of training and from there it decreased with semi-abrupt increase in training loss. Smaller learning rates were used but did not achieve the same validation results as our selected learning rate of $5 \times 10^{5}$. 

Validation results on the BERT model outperformed SpaCy in all three evaluation metrics from Table 4. The biggest performance gain was on the generalization of the BERT model trained on the FoodCombined to the held-out Cheklist test dataset. Here we see similar performance on the test set as we see with the validation set: an F1 score of 95.01\%, Recall of 95.20\%, and Precision score of 94.82\%. 

\begin{table}[H]
\centering
\begin{tabular}{|l|l|l|l|}
\hline
Model     & F1 score & Recall & Precision \\ \hline
SpaCy NER & 96.53    & 96.39  & 96.67     \\ \hline
BERT      & \textbf{98.55}    & \textbf{98.69}  & \textbf{98.40}     \\ \hline
\end{tabular}
\caption{Validation data set F1 Score, Recall and Precision}
\end{table}

\begin{table}[H]
\centering
\begin{tabular}{|l|l|l|l|}
\hline
Model     & F1 score & Recall & Precision \\ \hline
SpaCy NER & 55.97   & 49.20  &   64.91   \\ \hline
BERT      & \textbf{95.01}    & \textbf{95.20}  & \textbf{94.82}     \\ \hline
\end{tabular}
\caption{Test set data F1 Score, Recall, and Precision}
\end{table}

\subsection{Results: Examples}
In this section, we will examine and analyze some misclassifications done by our best performing model, BERT. The words highlighted in red are examples of false negatives and those in blue are the false positives. Orange words are misclassified as in they are labelled as one entity but there ground truth is a different non-empty entity. \\
\\
\textit{Example}: "\textcolor{blue}{Pierce pork} all over with a carving fork. Rub salt then liquid smoke over meat . Place roast in a slow cooker. Cover, and cook on Low for 16 to 20 hours, turning once during \textcolor{blue}{cooking time}. Remove meat from slow cooker , and shred , adding drippings".

The interesting point to note here is that our model misclassified "Pierce pork" as a single ingredient, whereas in the ground-truth the only ingredient present here is "pork". We believe this because of the data the model was trained. Some example of words that we noticed followed pork were, "ground pork", "grilled pork", "salted pork" etc, in all these the description followed by pork was the entire ingredient.\\
\\
\textit{Example}: "Heat honey and peanut butter together over low heat until creamy. Remove from heat and cool . Mix the rest of the ingredients into the \textcolor{red}{honey/peanut butter mixture} , except coconut. Shape into balls. Roll in coconut".

As you can see our model couldn't find the ingredient, which in this case was "honey/peanut butter mixture". We believe this is because our BERT model is accustomed to seeing ingredients which only consist of one-two tokens, sometimes three. So a four token ingredient would be an unseen use-case for our model. The token "honey/peanut" is also an interesting find because this is only specific to recipe data and would definitely not be found in  plain-text AI assistant queries. Another future consideration would to rid these token examples from the data all-together since it does not provide any useful information for our goal. \\
\\
\textit{Example}: "In a small bowl combine quick \textcolor{orange}{oats}, \textcolor{blue}{margarine}, and boiling water. cover and allow margarine to melt, and mixture to cool a bit. mix remaining \textcolor{blue}{cake} ingredients and mix well. spread into 9x13 greased and lightly floured cake pan . bake at 350f for approximately 35 minutes or until done . for icing : mix icing ingredients until well blended . spread over cooled cake and put under broiler until coconut".

This is an interesting example since our model labelled oats, margarine and cake as ingredients whereas in the ground-truth only oats was mentioned. In this case oats is incorrect since our model marked it as a `I-ING` thereby making it an intermediate token for an ingredient whereas the correct label was `B-ING', the beginning token of the ingredient. Cake is a simple misclassification. However, as we can observe margarine is a case where our model is in fact correct and our ground-truth is incorrect. 
\section{Conclusion \& Future Work}
In this report, we studied the task of extracting ingredients from raw text that is obtained from users looking to satisfy some kind of information need when it comes to food / recipes. To make train our models we combined and annotated datasets from various sources and to make them robust automated generation of real-world unseen data. We performed experiments and trained two different kinds of models: a traditional parser based one from SpaCy and a deep-learning transformer based model like BERT. \\
As we can see from Table 5,  BERT gives the best F1 score of 95.01. However, there are still improvements we can make. In future work, interested researchers can pursue the following. \\
\begin{enumerate}
    \item \textbf{Generate Noisy data}: We also looked at a paper in ASR \cite{larson2019telephonetic} to determine how we can generate noisy data similar to what it would look when generated by a text-to-speech model, to make our NER models ore robust. As mentioned earlier, The main application of this model will be in the TaskBot competition where the model will received text converted using ASR as its input. 
    \item \textbf{Generate data with more ingredients}: As mentioned, the usage of the Checklist framework didn't quite give us the kind of data we wanted to generate, A future direction to pursue would be building a testing framework similar to Checklist that can generate data for NER models and can handle multiple entities and their annotations. 
    \item \textbf{Use more types of tags}: We only tried to extract ingredients in this work, however it is not far-fetched to work on a model that can extract other aspects of a recipe along with ingredients. Some of these aspects could be: dietary restrictions, cooking utensils and cuisines. We have yet to come across a dataset that combines all of these since adding these annotations would require significant human effort. 
\end{enumerate}
\begin{scriptsize}
\bibliographystyle{IEEEtran}
\nocite{*}
\bibliography{ref.bib}

\begin{thebibliography}{10}
\providecommand{\url}[1]{#1}
\csname url@samestyle\endcsname
\providecommand{\newblock}{\relax}
\providecommand{\bibinfo}[2]{#2}
\providecommand{\BIBentrySTDinterwordspacing}{\spaceskip=0pt\relax}
\providecommand{\BIBentryALTinterwordstretchfactor}{4}
\providecommand{\BIBentryALTinterwordspacing}{\spaceskip=\fontdimen2\font plus
\BIBentryALTinterwordstretchfactor\fontdimen3\font minus
  \fontdimen4\font\relax}
\providecommand{\BIBforeignlanguage}[2]{{%
\expandafter\ifx\csname l@#1\endcsname\relax
\typeout{** WARNING: IEEEtran.bst: No hyphenation pattern has been}%
\typeout{** loaded for the language `#1'. Using the pattern for}%
\typeout{** the default language instead.}%
\else
\language=\csname l@#1\endcsname
\fi
#2}}
\providecommand{\BIBdecl}{\relax}
\BIBdecl

\bibitem{zhao2019extracting}
Y.~Zhao and J.~Handley, ``Extracting clinical concepts from user queries,''
  2019.

\bibitem{jain-pennacchiotti-2010-open}
\BIBentryALTinterwordspacing
A.~Jain and M.~Pennacchiotti, ``Open entity extraction from web search query
  logs,'' in \emph{Proceedings of the 23rd International Conference on
  Computational Linguistics (Coling 2010)}.\hskip 1em plus 0.5em minus
  0.4em\relax Beijing, China: Coling 2010 Organizing Committee, Aug. 2010, pp.
  510--518. [Online]. Available: \url{https://aclanthology.org/C10-1058}
\BIBentrySTDinterwordspacing

\bibitem{inproceedings}
G.~Popovski, S.~Kochev, B.~Seljak, and T.~Eftimov, ``Foodie: A rule-based
  named-entity recognition method for food information extraction,'' 02 2019.

\bibitem{IEfromrecipe}
R.~Agarwal and K.~Miller, ``Information extraction from recipes,''
  \url{https://nlp.stanford.edu/courses/cs224n/2011/reports/rahul1-kjmiller.pdf}.

\bibitem{10.1093/database/baz121}
\BIBentryALTinterwordspacing
G.~Popovski, B.~K. Seljak, and T.~Eftimov, ``{FoodBase corpus: a new resource
  of annotated food entities},'' \emph{Database}, vol. 2019, 11 2019, baz121.
  [Online]. Available: \url{https://doi.org/10.1093/database/baz121}
\BIBentrySTDinterwordspacing

\bibitem{LoperBird02}
E.~Loper and S.~Bird, ``{NLTK: The Natural Language Toolkit},'' in
  \emph{Proceedings of the ACL Workshop on Effective Tools and Methodologies
  for Teaching Natural Language Processing and Computational
  Linguistics}.\hskip 1em plus 0.5em minus 0.4em\relax Somerset, NJ:
  Association for Computational Linguistics, 2002, pp. 62--69,
  \url{http://arXiv.org/abs/cs/0205028}.

\bibitem{doccano}
\BIBentryALTinterwordspacing
H.~Nakayama, T.~Kubo, J.~Kamura, Y.~Taniguchi, and X.~Liang, ``{doccano}: Text
  annotation tool for human,'' 2018, software available from
  https://github.com/doccano/doccano. [Online]. Available:
  \url{https://github.com/doccano/doccano}
\BIBentrySTDinterwordspacing

\bibitem{ribeiro2020accuracy}
M.~T. Ribeiro, T.~Wu, C.~Guestrin, and S.~Singh, ``Beyond accuracy: Behavioral
  testing of nlp models with checklist,'' 2020.

\bibitem{spacy}
\BIBentryALTinterwordspacing
``{spaCy}: Model architectures.'' [Online]. Available:
  \url{https://spacy.io/api/architectures#parser}
\BIBentrySTDinterwordspacing

\bibitem{DBLP:journals/corr/abs-1810-04805}
\BIBentryALTinterwordspacing
J.~Devlin, M.~Chang, K.~Lee, and K.~Toutanova, ``{BERT:} pre-training of deep
  bidirectional transformers for language understanding,'' \emph{CoRR}, vol.
  abs/1810.04805, 2018. [Online]. Available:
  \url{http://arxiv.org/abs/1810.04805}
\BIBentrySTDinterwordspacing

\bibitem{transformer}
A.~Vaswani, N.~Shazeer, N.~Parmar, J.~Uszkoreit, L.~Jones, A.~N. Gomez,
  L.~Kaiser, and I.~Polosukhin, ``Attention is all you need,''
  \url{https://arxiv.org/pdf/1706.03762.pdf}.

\bibitem{larson2019telephonetic}
C.~Larson, T.~Lahlou, D.~Mingels, Z.~Kulis, and E.~Mueller, ``Telephonetic:
  Making neural language models robust to asr and semantic noise,'' 2019.

\bibitem{8995569}
G.~Popovski, B.~K. Seljak, and T.~Eftimov, ``A survey of named-entity
  recognition methods for food information extraction,'' \emph{IEEE Access},
  vol.~8, pp. 31\,586--31\,594, 2020.

\bibitem{li2020survey}
J.~Li, A.~Sun, J.~Han, and C.~Li, ``A survey on deep learning for named entity
  recognition,'' 2020.

\end{thebibliography}
\end{scriptsize}
\end{document}